\documentclass[10pt,twocolumn,letterpaper]{article}

\usepackage[pagenumbers]{cvpr} 

\usepackage{graphicx}
\usepackage{amsmath}
\usepackage{amssymb}
\usepackage{booktabs}
\usepackage{makecell}
\usepackage{multirow}
\usepackage{arydshln}

\usepackage[pagebackref,breaklinks,colorlinks]{hyperref}

\usepackage[capitalize]{cleveref}
\crefname{section}{Sec.}{Secs.}
\Crefname{section}{Section}{Sections}
\Crefname{table}{Table}{Tables}
\crefname{table}{Tab.}{Tabs.}

\def\cvprPaperID{5301} 
\def\confName{CVPR}
\def\confYear{2022}

\begin{document}

\title{General Facial Representation Learning in a Visual-Linguistic Manner}

\renewcommand{\thefootnote}{\fnsymbol{footnote}}

\author{Yinglin Zheng\textsuperscript{1}\footnotemark[1]~~\footnotemark[2]~~~Hao Yang\textsuperscript{2}\footnotemark[1]~~~Ting Zhang\textsuperscript{2}~~~Jianmin Bao\textsuperscript{2}~~~Dongdong Chen\textsuperscript{3}\\Yangyu Huang\textsuperscript{2}~~~Lu Yuan\textsuperscript{3}~~~Dong Chen\textsuperscript{2}~~~Ming Zeng\textsuperscript{1}~~~Fang Wen\textsuperscript{2}\\
{\textsuperscript{1}School of Informatics, Xiamen Unversity}\\
{\textsuperscript{2}Microsoft Research Asia}~~~{\textsuperscript{3}Microsoft Cloud+AI}\\
{\tt\small \{zhengyinglin@stu., zengming@\}xmu.edu.cn, cddlyf@gmail.com,}
\\
{\tt\small \{haya, tinzhan, jianbao, yangyu.huang, luyuan, doch, fangwen\}@microsoft.com}
\\
{\small\url{https://github.com/faceperceiver/farl}}
}

\maketitle

\footnotetext[1]{Equal contribution.}
\footnotetext[2]{Work done during internship at MSRA.}
\renewcommand{\thefootnote}{\arabic{footnote}}

\begin{abstract}   
How to learn a universal facial representation that boosts all face analysis tasks?
This paper takes one step toward this goal. 
In this paper, we study the transfer performance of pre-trained models on face analysis tasks and introduce a framework, called \textbf{FaRL}, for general facial representation learning. 
On one hand, the framework involves a contrastive loss to learn high-level semantic meaning from image-text pairs.
On the other hand, we propose exploring low-level information simultaneously to further enhance the face representation by adding a masked image modeling.
We perform pre-training on LAION-FACE, a dataset containing a large amount of face image-text pairs, and evaluate the representation capability on multiple downstream tasks.
We show that FaRL achieves better transfer performance compared with previous pre-trained models. 
We also verify its superiority in the low-data regime. 
More importantly, our model surpasses the state-of-the-art methods on face analysis tasks including face parsing and face alignment.
\end{abstract}
\section{Introduction}
\label{sec:intro}

Face analysis tasks are of crucial importance to social interactions 
and have received extensive attention over the past decades.
Many existing state-of-the-art results~\cite{te2021agrnet,kumar2020luvli,cao2018partially} come from   
deep neural networks with supervised learning.
However, such supervised models,
in order to learn appropriate feature representations for each given task,
are studied separately with large-scale manually annotated data which is expensive and difficult to acquire, especially for some face tasks such as face parsing and face alignment.

Recently, visual representation learning in computer vision appears to be paved by a popular learning paradigm, pre-training, due to the remarkable success of the groundbreaking models in Natural Language Processing such as BERT\cite{devlin2018bert} and GPT-series~\cite{radford2018improving,radford2019language,brown2020language}, followed by a wide variety of multiple techniques~\cite{peters2018deep,lewis2019bart, raffel2019exploring,joshi2020spanbert,dong2019unified }.
Thereafter in vision, 
many attempts~\cite{he2020momentum,grill2020bootstrap,chen2020simple} along this path have been proposed,
showing promising results using approaches related to contrastive loss~\cite{hadsell2006dimensionality,dosovitskiy2014discriminative,oord2018representation,bachman2019learning,caron2020unsupervised}.

Meanwhile in the area of vision-language tasks involving multi-modality,
there are studies~\cite{lu2019vilbert,su2019vl,li2020unicoder,radford2021learning,jia2021scaling,li2021align} exploring learning directly from large, freely available
image-text pairs up to hundreds of millions. 
Their results show that natural language supervision is beneficial for visual representation learning, bringing superior performance to most tasks on general images.
Such pre-training has several advantages: 1) showing promising few-shot transfer performance, alleviating the hard-acquired labeled data issue; 2) enabling convenient deployment by extracting general feature representation once and then applying to diverse downstream tasks.
Yet when it comes to the face domain, one of the most important domains in computer vision, the effectiveness of pre-training is relatively unexplored.

In this paper, we study the behaviors of transfer performance of pre-trained models on face analysis tasks
and introduce a framework, FaRL, to learn general facial representation in a visual-linguistic manner.
Instead of crawling images and texts from the Web by manually designed face-related queries, 
we create a dataset by filtering out from a large openly available image-text-pair dataset \cite{schuhmann2021laion}, resulting in a subdataset containing 20 million face images, denoted as LAION-FACE. 

We adopt the widely used contrastive loss to pull the embeddings of matched image-text pair
together while pushing those of non-matched image-text pairs apart, which provides high-level semantic meaning.
We propose exploring supplementary low-level information simultaneously to
further enhance the face representation, by adding a masked image modeling inspired from BEiT\cite{bao2021beit}.
We perform pre-training on LAION-FACE and evaluate the representation capability with the frozen backbone, as our ultimate goal is to offer a general facial representation that can be quickly adapted to downstream tasks.

We measure the performance on 
several important downstream face tasks, including face parsing, face alignment and face attribute prediction, for which labeled data is usually difficult to acquire.
We show that better transfer performance can be achieved compared with other pre-trained models.
We also demonstrate its superiority in the low-data regime.
Moreover, our model outperforms the state-of-the-art methods on face tasks including face parsing and face alignment.

In summary, our major contributions are as follows:
\begin{enumerate}
	\item We present an extensive study about the transferable visual models learned in a visual-linguistic manner on versatile face analysis tasks, which is relatively unexplored in the literature.
	\item We introduce a new framework, exploring low-level and high-level information simultaneously for better representation. We achieve superior transfer performance than previous pre-training approaches, and more importantly on face parsing and face alignment, our model has surpassed the state-of-the-art methods.
\end{enumerate}

\section{Related Works}
\label{sec:related_works}
\subsection{Visual Representation Learning}
Ever since the pioneering work~\cite{krizhevsky2012imagenet} on ImageNet recognition followed by
numerous improvements~\cite{simonyan2014very,he2016deep,huang2017densely,xie2017aggregated},
ImageNet classification~\cite{deng2009imagenet} has been the de facto pre-training task for visual representation learning.
The backbones for
a wide range of visual tasks including image classification, object detection, semantic segmentation and human pose estimation are often initialized from ImageNet pre-trained weights with the goal of less task-specific data and fewer training epochs.
Supervised pre-training then becomes the predominant recipe in visual representation learning, with a scaling-up trend in the size and complexity of the network~\cite{kolesnikov2020big} as well as the size of the training dataset, \eg, JFT-300M~\cite{sun2017revisiting} and Instagram-1B~\cite{mahajan2018exploring}.

As pre-training with large Transformer-based networks on large corpora~\cite{devlin2018bert,radford2018improving,radford2019language,brown2020language,lewis2019bart, dong2019unified} has pushed the state of the art forward in natural language processing,
there emerge lots of inspired advancements in visual representation learning.
The vision Transformer~\cite{dosovitskiy2020image, touvron2021training} applied a standard Transformer directly to image classification by splitting an image into patches, similar to the tokens in NLP.
iGPT~\cite{chen2020generative} defined the auto-regressive and BERT objective in the context of images by predicting pixels whose color is indexed by k-means clustering.
On the other hand, BEiT~\cite{bao2021beit} designed a BERT-style loss based on the visual discrete tokens obtained by the discrete variational auto-encoder.

Subsequently, the remarkable success of natural language processing also flourished the visual-language pre-training, where large amounts of freely available image text pairs can be acquired on the Internet.
Without relying on the the pre-trained object detector to extract image region features as previous works~\cite{tan2019lxmert,chen2020uniter,li2020oscar,huang2021seeing,zhang2021vinvl,yu2020ernie},
CLIP~\cite{radford2021learning} and ALIGN~\cite{jia2021scaling} adopt the
contrastive loss, an effective loss in self-supervised representation learning, to pull the embedding of matched image-text pairs together while pushing those
of non-matched pairs apart, followed by ALBEF~\cite{li2021align} which further improves the visual-language pre-training. 

Another line of works~\cite{he2020momentum,chen2020improved,chen2021empirical,grill2020bootstrap,chen2020simple,chen2020big} focus on learning visual representation without any supervision.
Among the most successful of recent efforts,
the critical core also relates to contrastive loss, measuring the similarities of augmented image pairs in a representation space.

\subsection{Facial Representation Learning}
In the area of face analysis, most tasks \cite{lin2019face,kumar2020luvli,te2021agrnet,te2020edge,huang2021adnet,cao2018partially} are solved by supervised training 
with manually labeled data.
Such supervised methods require a large number of training samples and may suffer from overfitting because of the massive number of model parameters.
While pre-training has shown impressive performance on few-shot learning and also help reduce overfitting~\cite{hendrycks2019using},
the effectiveness of pre-training on face domain is rarely explored yet.
Another major advantage of pre-training is that there will be a universal facial representation that can be well transferred to a variety of downstream tasks, which is particularly desirable for
resource-limited mobile devices.

There exist several works addressing few-shot learning~\cite{browatzki20203fabrec,shu2021learning} and transfer learning~\cite{aneja2020generalized,zhuang2021few} on a specific face task instead of examining pre-training over diverse face tasks.
Closely related to our work is~\cite{bulat2021pre}, which explored unsupervised pre-training for facial representation learning.
In contrast, we bring clarity to an unexplored regime, weakly-supervised pre-training, on face domain
as it has been shown in~\cite{radford2021learning} that leveraging 
massive web image-text pairs, which provides weak supervision for images, is helpful
for learning visual representations in few-shot scenarios.

\section{FaRL}
\label{sec:face_clip}

\subsection{Visual Linguistic Face Data}
Our goal is to learn a transferable facial representation in a visual-linguistic manner. 
For this purpose, we start by collecting a sufficiently large dataset that contains image-text pairs where the image includes a face region and the text label is natural language.
Existing face datasets~\cite{guo2016msceleb,LFWTech,liu2020new,sagonas2016300,lee2020maskgan,liu2015deep} are mainly designed for a specific face task.
For example,
the associated labels of corresponding datasets are identities, semantic masks, landmark locations, attribute tags for face recognition, face parsing, face alignment and attribute prediction respectively.

\begin{figure}[t]
  \centering
  \includegraphics[width=\linewidth]{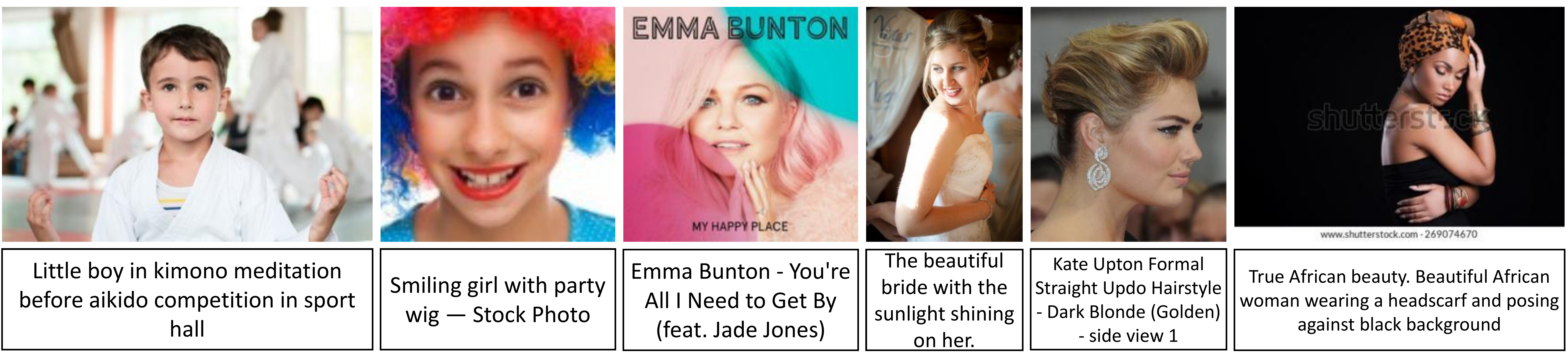}
   \caption{Image-text pairs randomly sampled from LAION-FACE. Web texts are not always accurate but often easier to acquire. 
   }
   \label{fig:laion-face}
  \vspace{-0.8em}
\end{figure}

\begin{figure}[t]
\vspace{-0.5em}
  \centering
  \includegraphics[width=.95\linewidth]{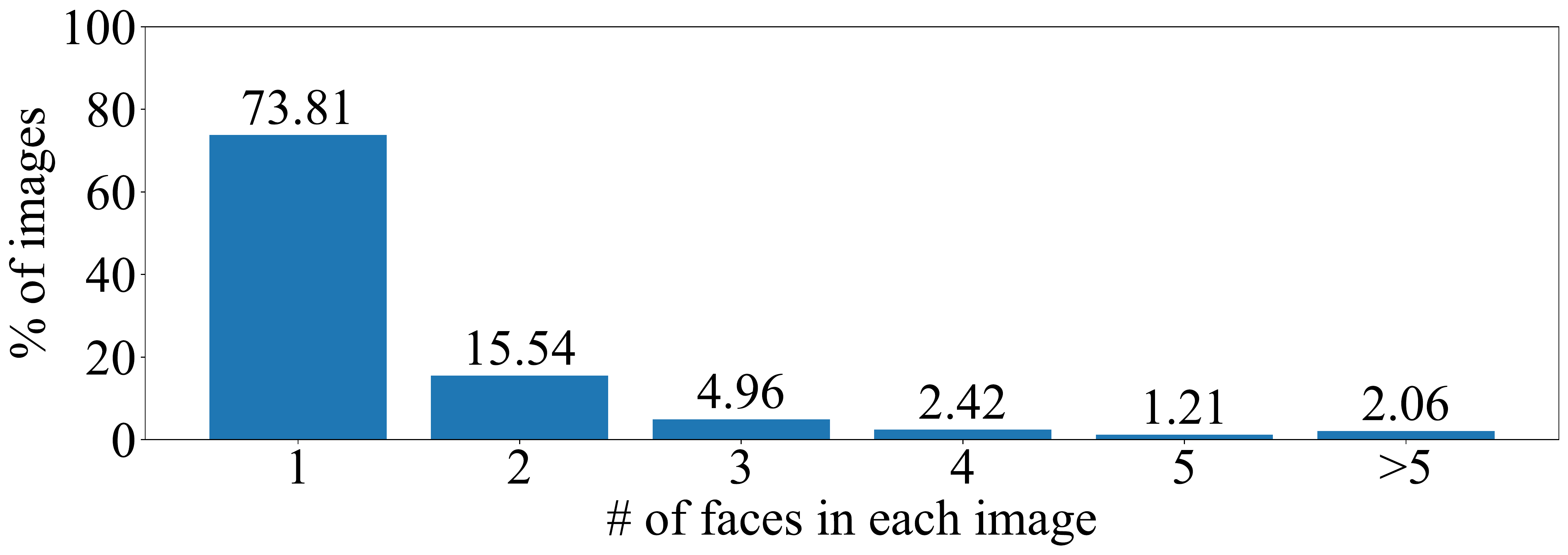}
   \caption{Distribution of \#faces in each image in LAION-FACE. }
   \label{fig:face-dist}
\end{figure}

To enable learning from natural language supervision, which benefits from the large quantities of data available on the Internet,
we construct a new dataset consisting of 20 million image-text pairs.  
Specifically, we leverage the large openly available image-text-pair dataset, LAION~\cite{schuhmann2021laion}, that contains 400 million samples.
To filter out those non-face images,
we adopt a face detector, RetinaFace~\cite{deng2019retinaface}, to identify the presence of a face in an image.
20 million pairs are randomly sampled from those whose face detection scores are greater than 0.9. The resulting dataset is denoted as LAION-FACE.
Figure~\ref{fig:laion-face} shows some image-text-pair samples from the dataset. The distribution of the number of faces in each image is shown in Figure~\ref{fig:face-dist}.

\subsection{Image-text Contrastive Learning}
Following~\cite{radford2021learning,jia2021scaling}, we adopt the image-text contrastive loss, which
has been shown to be more compute-efficient than generative models~\cite{chen2020generative} 
and learn better representation than the predictive counterpart~\cite{tian2020contrastive}.
The contrastive learning learns by comparing,
according to some notion of similarity.
Precisely, consider a given image-text pair $\{T,I\}$, 
the extracted feature representations are $\{f^I_{cls}, f^I_1, \cdots, f^I_N\} = E_I(I), \{f^T_{eos},f^T_1, \cdots, f^T_M\} = E_T(T)$,
where $E_I$ denotes the image Transformer-based encoder and $E_T$ denotes the text Transformer-based encoder.
$cls$ is short for class token, $eos$ is short for end of sequence token and $1,\cdots,N(M)$ denotes the index of visual (language) tokens.
The features from the $cls$ ($eos$) token are then fed into a projection head (a small MLP) to obtain the metric embeddings,
\ie $e^I=P_I(f^I_{cls}), e^T = P_T(f^T_{eos})$. 
The contrastive loss, in the scenario of image-text pairs, is given as
\begin{align}
	L_I &= - \frac 1 B \sum_{i=1}^B \log \frac {\exp(e_i^Ie_i^T/\sigma)} {\sum_{j=1}^B \exp(e_i^Ie_j^T/\sigma)},\nonumber\\
	L_T &= - \frac 1 B \sum_{i=1}^B \log \frac {\exp(e_i^Te_i^I/\sigma)} {\sum_{j=1}^B \exp(e_i^Te_j^I/\sigma)},
  \label{equ:itc_loss}
\end{align}
where $B$ is the number of image-text pairs in a mini-batch, and $\sigma$ is the temperature 
to scale the logits, which is learned together with all other parameters.

\begin{figure}[t]
  \centering
  \includegraphics[width=\linewidth]{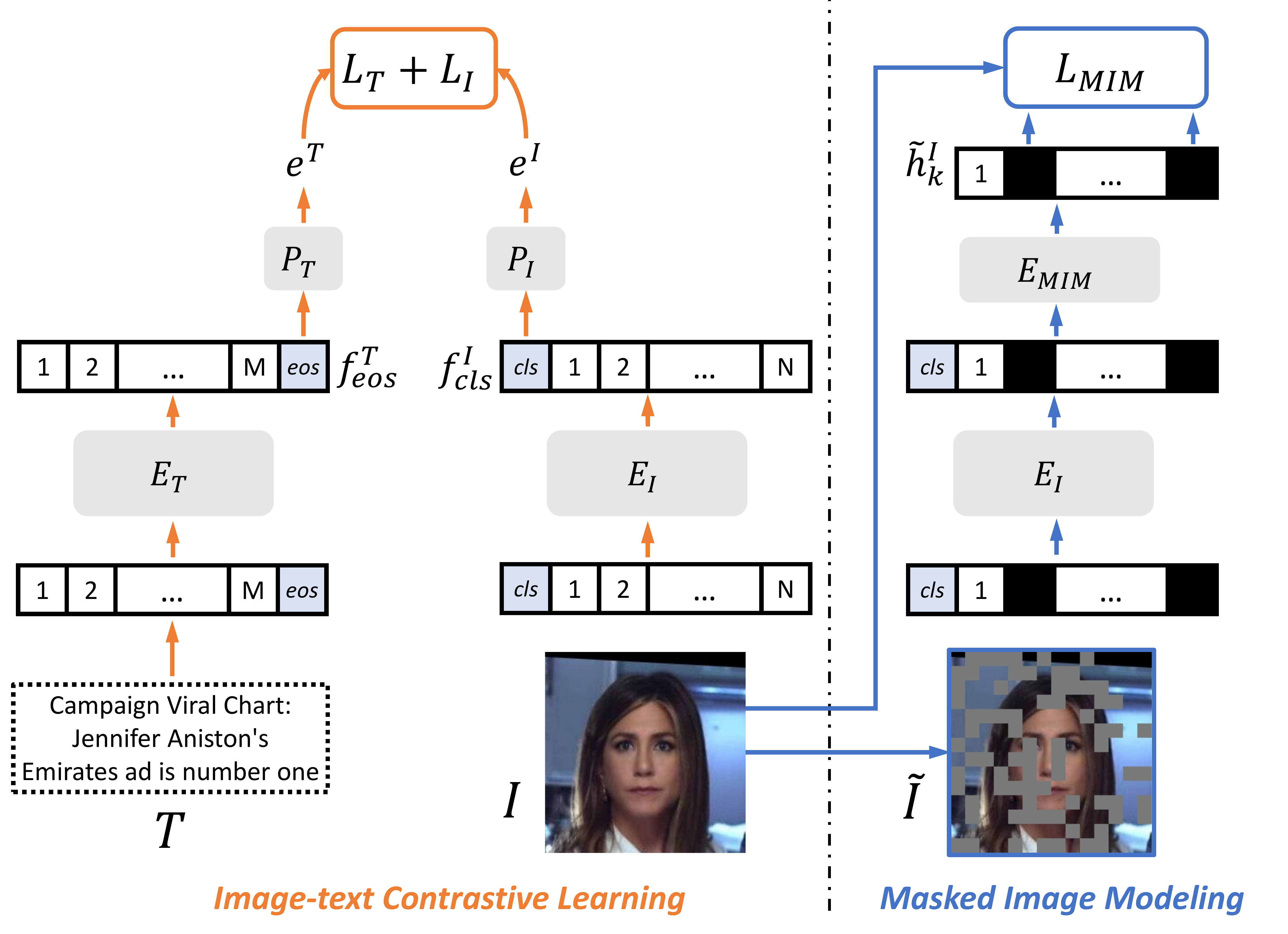}
   \caption{Illustrating our pre-training framework. We integrate masked image modeling with image-text contrastive learning. The two $E_I$ in this figure stand for the same image encoder. After the pre-training, we use $E_I$ to boost downstream face tasks.}
   \label{fig:framework}  
\end{figure}

\subsection{Masked Image Modeling}
It is intuitively plausible that
the image-text contrastive learning facilitates to learn semantic feature representations from text
about concrete or visualizable concepts.
To further enhance the face representation, we add a masked image modeling task that masks some image patches in the input and predicts the visual tokens corresponding to the masked patches.
This objective is similar to image inpainting aiming at filling in holes of an image, which is a representative low-level vision task.
We hypothesize that this masked image modeling will help the features to capture low-level information, providing complementary information to high-level semantics.

Formally, let $\tilde{I}$ be the masked image, where some image patches are randomly masked.
That is to say, if a given input image $I$ is split into $N$ image patches $\{I_1,\cdots,I_N\}$, the masked image $\tilde{I}$ is also represented as $N$ image patches $\{\tilde{I}_1, \cdots, \tilde{I}_N\}$ with 
\begin{align}
	\tilde{I}_k =  \left\{ 
	\begin{array}{cc}
		I_k, & k \notin \mathcal{M}\\
		m, & k \in \mathcal{M}\\
	\end{array}
	\right. ,
\end{align}
where $\mathcal{M}\subset\{1,\cdots,N\}$ denotes the positions where image patches are masked, and $m$ is the masked token, a learnable vector as same dimension as non-masked patches.
After we get the features from the image encoder, $\{\tilde{f}^I_{cls}, \tilde{f}^I_1, \cdots, \tilde{f}^I_N\} = E_I(\tilde{I})$, 
we feed them into a small Transformer which outputs the final hidden vectors, $\{\tilde{h}^I_{cls}, \tilde{h}^I_1, \cdots, \tilde{h}^I_N\} = E_\textit{MIM}(\tilde{f}^I_{cls}, \tilde{f}^I_1, \cdots, \tilde{f}^I_N)$.
The objective is to predict the masked region from the corresponding hidden vectors
$\{\tilde{h}^I_k, k \in \mathcal{M}\}$.
Instead of directly predicting pixels requiring huge memory consumption, 
the discrete variational autoencoder~\cite{oord2017neural} is utilized here to first encode each image patch to 
one of $|\mathcal{V}|$ possible values, with $\mathcal{V}$ being the vocabulary of the autoencoder.
Thereafter, a classification layer is attached on the hidden vector $\tilde{h}^I_k$ to predict
the corresponding masked patch's index among $\{1, \cdots, |\mathcal{V}|\}$.
The loss function is given as,
\begin{align}
	L_\textit{MIM} = -\sum_{k \in \mathcal{M}}\log p\left(q^k_{\phi}(I)|\tilde{I}\right),
\end{align}
where $p(q^k_{\phi}(I)|\tilde{I})$ denotes the classification score of classifying the $k$-th
hidden vector belonging to the visual token $q_{\phi}^k(I)$,
where $q_{\phi}$ is the categorical distribution.

The whole framework is illustrated in Figure \ref{fig:framework}.
In the experiment, we directly use the publicly available discrete variational autoencoder described in~\cite{ramesh2021zero}.

\subsection{Pre-training Details}
\label{ssec:pretraining_details}

\noindent
\textbf{Model architecture.}
Our model consists of an image encoder $E_I$, a text encoder $E_T$ and a masked image modeling module $E_\textit{MIM}$.
We implement the image encoder $E_I$ following prior works~\cite{dosovitskiy2020image,bao2021beit,radford2021learning} for fair comparison.
Specifically, it is a 12-layer 768-width visual Transformer ViT-B/16~\cite{dosovitskiy2020image} with 87M parameters and $224\times 224$ input.
The input image is first split into $14\times 14$ image patches, followed by a linear projection to obtain $14\times 14$ patch embeddings.
A learnable \emph{cls} token is prepended to these 196 embeddings, which is then added by 197 positional embeddings.
We design the text encoder $E_T$ following \cite{radford2021learning}.
It is a 12-layer, 512-width and 8-head Transformer with 63M parameters.  
We fix the number of input text tokens to 77, and perform truncation or padding if the input length does not match.
Finally we project features from both the image \emph{cls} token and the text \emph{eos} token to 512 dimension to calculate the contrastive loss.
The masked image modeling module $E_\textit{MIM}$ is implemented as a one-layer Transformer for both simplicity and performance consideration.

\noindent
\textbf{Pre-training setup.}
We train our model from scratch with random initialized weights.
We run pre-training for 16 epochs with batch-size 5120 on 32 Nvidia V100 GPUs.
The AdamW~\cite{loshchilov2017decoupled} optimizer is adopted with a weight decay set as 0.05.
The learning rate is initialized as 1e-6, which is warmed up to 1e-3 in one epoch and then cosine decay to 9e-4 in 15 epochs.
The softmax temperature $\sigma$ is initialized as $0.07$.
The input face image is aligned to the mean face as input. For images containing more than one face, we will choose one randomly. 
During pre-training, every input image will be fed into image encoder twice: one for image-text contrastive learning, one with randomly masked (at most 75 patches) image patches for masked image modeling.

\subsection{Downstream Face Tasks}

We adapt our model to multiple downstream face tasks that span various categories (segmentation, regression and classification) to evaluate its transfer performance:

\noindent\textbf{Face parsing} predicts pixel-wise regions of face components. 
Two popular datasets are used for this task: LaPa \cite{liu2020new} and CelebAMask-HQ \cite{lee2020maskgan}. LaPa contains over 22K images, 18,176 for training and 2K for testing, each annotated with an 11-category pixel-level label map. CelebAMask-HQ consists of around 30K facial images, 24,183 for training and 2,824 for testing, each annotated with a 19-category label map including not only facial components but also body parts and accessories like eyeglasses, earrings and necklaces. Following \cite{liu2020new,te2020edge,te2021agrnet},  
the F1 scores of facial components are used to measure the performance.

\noindent\textbf{Face alignment} aims to regress 2D face landmark coordinates on a face image. We use three popular datasets: AFLW-19 \cite{zhu2016unconstrained} with 20K images for training and 4,386 for testing, each annotated with 19 landmarks; 300W \cite{sagonas2016300,sagonas2013semi,sagonas2013300} with 3,837 images for training and 600 for testing, each annotated with 68 landmarks; WFLW \cite{wu2018look} with 7,500 training data and 2,500 testing data, each annotated with 98 landmarks. Following common practice, we use normalized mean error (NME), failure rate (FR) and AUC as the metric.

\noindent\textbf{Face attributes recognition} predicts multiple attributes (\eg gender, age and race) of a given face image, which can be viewed as a multi-label classification. Two datasets are adopted: CelebA \cite{liu2015deep} and LFW \cite{liu2015deep}. CelebA consists of over 202K facial images, while LFWA consists of 13,143 images, both with 40 attribute annotations per image. Following \cite{liu2015deep,shu2021learning}, on CelebA, we use 162,770 for training and 19,962 for testing; on LFWA, We use 6,263 for training and the rest for testing. The averaged accuracy over all attributes is used as the metric.

\section{Experiments}
\label{sec:experiments}

\begin{table*}[h]
\centering
\footnotesize
\begin{tabular}{cccc|c|c|c}

\multicolumn{4}{c|}{Pre-training Settings} & \multicolumn{3}{c}{Downstream Performances} \\
\hline

\makecell{Dataset} & \makecell{Scale} & \makecell{Supervision\\Source} & \makecell{Method} & 
\makecell{LaPa \\ F1-mean$\uparrow$} & \makecell{AFLW-19 \\ $\mathrm{NME}_\mathrm{diag} \downarrow$} & \makecell{CelebA \\ mAcc $\uparrow$} \\

\hline

ImageNet-1K & 1.3M & images & MoCo v3 \cite{chen2021empirical}                            & 91.86 & 1.007 &  90.23  \\
ImageNet-22K & 14M & images & BEiT \cite{bao2021beit}                                     & 91.29 & 1.076 &  89.71  \\
ImageNet-22K & 14M & images + human labels & ViT \cite{dosovitskiy2020image}              & 91.61 & 1.004 &  90.77  \\
ImageNet-1K & 1.3M & images + human labels & DeiT \cite{touvron2021training}              & 92.00 & 1.003 &  89.79  \\
WIT & 400M & images + web text & CLIP \cite{radford2021learning}                          & 92.21 & 0.995 & 90.86   \\
MS1MV3 & 5.1M & face images + human labels & Face Transformer \cite{zhong2021face}    & 91.09 & 1.031 & 90.77   \\

\hline 

\multirow{1}{*}{\makecell{LAION-FACE}} & \multirow{1}{*}{\makecell{20M}}
 & face images + web text & FaRL (Ours)           & \textbf{92.32}  & \textbf{0.991}  &  \textbf{91.39} \\

\end{tabular}
\caption{Comparing the universal capability of pretrained vision transformers on multiple downstream face tasks, including face parsing (LaPa), face alignment (AFLW-19) and face attributes recognition (CelebA). All pre-trained backbones are frozen with only heads fine-tuned on downstream data. }
\label{tab:compare_pretrains}
\end{table*}

\begin{table*}[t]
\footnotesize
\centering
\begin{subtable}{0.32\textwidth}
\centering
\begin{tabular}{c|ccccc}
{Pre-trained Models} & $1\%$ & $10\%$ & $100\%$  \\
\hline
MoCo v3 \cite{chen2021empirical} & 86.47 & 90.18 & 91.86 \\
BEiT \cite{bao2021beit} & 85.01 & 89.21 & 91.29 \\
ViT \cite{dosovitskiy2020image} & 86.64 & 89.97 & 91.61 \\
DeiT \cite{touvron2021training} & 87.24 & 90.45 & 92.00 \\
CLIP \cite{radford2021learning} & \underline{88.13} & \textbf{90.91} & \underline{92.21} \\
{\tiny FaceTransformer} \cite{zhong2021face} & 86.42 & 89.54 & 91.09 \\
\hline
FaRL &  \textbf{88.21} & \textbf{90.91} & \textbf{92.32} \\
\end{tabular}
\caption{LaPa (F1-mean $\uparrow$).}
\end{subtable}
\hfill%
\begin{subtable}{0.32\textwidth}
\centering
\begin{tabular}{c|ccccc}
{Pre-trained Models} & $1\%$ & $10\%$ & $100\%$  \\
\hline
MoCo v3 \cite{chen2021empirical} & 1.41 & 1.19 & 1.007 \\
BEiT \cite{bao2021beit} & 1.94 & 1.37 & 1.076 \\
ViT \cite{dosovitskiy2020image} & 1.37 & 1.16 & 1.004 \\
DeiT \cite{touvron2021training} & 1.41 & 1.17 & 1.003 \\
CLIP \cite{radford2021learning} & \textbf{1.30} & \textbf{1.11} & \underline{0.995} \\
{\tiny FaceTransformer} \cite{zhong2021face} & 1.41 & 1.21 & 1.031 \\
\hline
FaRL & \underline{1.35} & \underline{1.15} & \textbf{0.991} \\
\end{tabular}
\caption{AFLW-19 ($\mathrm{NME}_{\mathrm{diag}} \downarrow$)}
\end{subtable}
\hfill%
\begin{subtable}{0.32\textwidth}
\centering
\begin{tabular}{c|ccccc}
{Pre-trained Models} & $1\%$ & $10\%$ & $100\%$  \\
\hline
MoCo v3 \cite{chen2021empirical} &87.85 & 89.76 & 90.23\\
BEiT \cite{bao2021beit} & 85.64 & 88.74 & 89.71\\
ViT \cite{dosovitskiy2020image} & \underline{89.20} & 90.21 & 90.77\\
DeiT \cite{touvron2021training} & 86.73 & 89.00 & 89.79\\
CLIP \cite{radford2021learning} & 89.09 & \underline{90.47} & \underline{90.86}\\
{\tiny FaceTransformer} \cite{zhong2021face} & 87.42 & 90.32 & 90.77\\
\hline
FaRL & \textbf{89.66}& \textbf{90.99} &  \textbf{91.39}\\
\end{tabular}
\caption{CelebA (mAcc $\uparrow$).}
\end{subtable}
\caption{Comparing the generality of pretrained vision transformers w.r.t few-shot settings. We randomly sample subsets of the downstream training sets and evaluate the performances on the same test sets. All pre-trained backbones are frozen without fine-tuning on downstream data. Best performances are shown in bold and the second ones underlined.}
\label{tab:fewshot}
\end{table*}

\subsection{Setup}
Different heads are designed for different downstream tasks. All the heads exploit features not only from the last layer of the visual Transformer $E_I$, but also from some intermediate layers of it. Let $\{f^{I}_{\textit{cls},k}, f^{I}_{1,k}, f^{I}_{2,k}, \cdots, f^{I}_{N,k}\}$ be the feature representations from the $k$-th layer, with $1\leq k \leq 12$, since $E_I$ consists of 12 layers in total.

We design a simple head for face attributes recognition. Let $\mathcal{K}$ be the set of selected layers for downstream. We compute three feature vectors from each layer $k\in\mathcal{K}$: the $cls$ token feature $f^I_{\textit{cls},k}$, the mean of all non-$cls$ token features and the global max-pooling of all non-$cls$ token features. These $3\times|\mathcal{K}|$ vectors are then layer-normalized and linearly combined into one vector through learnable weights,
appended with a fully connected layer to generate the logits for multi-label binary classification. The head is trained with binary-cross-entropy loss and AdamW \cite{loshchilov2017decoupled} optimizer. We adopt a learning rate of $0.3$ and cosinely decay to zero in 100 epochs.  

Face parsing requires spatial distributions from the $N$ non-$\textit{cls}$ tokens on each layer. Note these non-$\textit{cls}$ tokens corresponds to image patches, hence they can be reshaped to a 2D feature map of size $\sqrt{N}\times\sqrt{N}$ ($14\times 14$ in our case). Following \cite{bao2021beit,liu2021swin}, UperNet \cite{xiao2018unified} is used to integrate multi-layer feature maps to generate a final feature map. A $1\times 1$ conv is appended to compute the parsing logits. The head is trained with cross-entropy loss, using AdamW with a learning rate 1e-3 and weight decay 1e-5. Tanh-warping \cite{lin2019face} is employed to balance the segmentation performance between the inner facial components and the hair region.

The face alignment head predicts heatmaps of the 2D landmark points, as is practiced by \cite{wang2019adaptive,kumar2020luvli,huang2021adnet}. We render groundtruth landmark points as Gaussian heatmaps of size $128\times128$ with a one-pixel $\sigma$ and values $\in [0, 1]$. Instead of using those complex loss functions designed by \cite{wang2019adaptive,huang2021adnet,feng2018wing}, we simply train the head with a soft-label cross-entropy loss. UperNet \cite{xiao2018unified} is also employed to output the heatmap logits. We train the head using AdamW with learning rate $0.01$ and weight decay 1e-5.

For all downstream tasks, we follow \cite{bao2021beit} and set the selected layers $\mathcal{K}=\{4, 6, 8, 12\}$. The above setups are also adopted for other pre-trained models for fair comparison. Please refer to the appendix for more details.

\subsection{Comparing with Pre-trained Transformers}
\label{ssec:fixbb}

We are curious about a question: \emph{given one face image as input, is it possible that the  feature output from our pretrained model could be quickly adapted to benefit all downstream tasks?}
To answer this question, we freeze our pretrained model, extract features from input face images using the same frozen encoder $E_I$, and directly leverage the output features to facilitate downstream training.

We compare FaRL with other publicly available pretrained Transformers. For fairness, all models share the same backbone structure (ViT-B/16)\footnote{Except Face Transformer, which has smaller patch size of 8 but a larger transformer layer number of 20. We set $\mathcal{K}=\{6,9,13,20\}$ for Face Transformer.}. We ensure that they only differ in backbone weights, with all other settings (head structures, training hyper-parameters \etc) identical.

We compare with six pre-trainings: 1) MoCo v3 \cite{chen2021empirical}, a model pretrained on ImageNet-1K using image-wise contrastive learning; 2) BEiT \cite{chen2021empirical}, pretrained on ImageNet-22K with mask image modeling only; 3) ViT \cite{dosovitskiy2020image}, the classic vision transformer pretrained with large-scale human annotated labels from ImageNet-22K, in a fully supervised manner; 4) DeiT \cite{touvron2021training}, an improved version of ViT that is pretrained on ImageNet-1K with distillation, also in a fully supervised manner; 5) CLIP \cite{radford2021learning}, which is pretrained on 400M visual-linguistic data with image-text contrastive loss only; 6) Face Transformer \cite{zhong2021face}, a transformer pretrained on a large face recognition dataset MS1MV3\cite{deng2019lightweight,guo2016msceleb,deng2019arcface} with human annotated face identity labels.

We illustrate their results on three downstream tasks in Table \ref{tab:compare_pretrains}: LaPa for face parsing, AFLW-19 for face alignment and CelebA for face attributes recognition. FaRL exhibits consistent advantage over other pre-trainings on all tasks. The weakly supervised FaRL outperforms not only self-supervised pre-trainings (MoCo v3, BEiT)
but also fully supervised pre-trainings (DeiT, ViT).
FaRL also surpasses CLIP, a weakly-supervised model pre-trained with a significantly larger data scale (400M image-text pairs).
Both pre-trained on face data, FaRL also excels the fully supervised Face Transformer by a large margin.

We also conduct evaluations under few-shot settings with limited downstream training data. Results reported in Table \ref{tab:fewshot} show that
FaRL achieves the best few-shot performances on both face parsing (LaPa) and face attributes recognition (CelebA). However, with $1\%$ and $10\%$ downstream training data, FaRL is outperformed by CLIP on face alignment (AFLW-19). We conjecture that the good generalizability of CLIP might come from their private 400M pre-training data: with an extremely large data scale (20 times the size of LAION-FACE), their dataset should involve a significant number of faces no less than LAION-FACE.

\subsection{Ablating Pre-training Components}
\label{sec:ablation}
\begin{table}

\centering
\scriptsize
\setlength\tabcolsep{2pt}
\begin{tabular}{l|ccc}
\makecell{Pre-training Settings} 

& \makecell{LaPa\\F1-mean$\uparrow$} & \makecell{AFLW-19\\$\mathrm{NME}_\mathrm{diag}\downarrow$} 
& \makecell{CelebA\\mAcc$\uparrow$} \\
\hline 
ITC                                                      & 91.75 & 1.009  & 91.31\\
ITC+MIM\textsubscript{1}                             & 91.82 & 1.004    & 91.22   \\
\emph{ITC+MIM\textsubscript{1}+ALIGN} (FaRL)         & \textbf{92.32} & \textbf{0.991}   & \underline{91.39}  \\
ITC+MIM\textsubscript{6}+ALIGN                       & \underline{92.19} &  1.002   & 91.38   \\
ITC+MIM\textsubscript{6}                               & 91.99 & \underline{0.992}   & 91.20  \\
ITC+ALIGN                                              & 91.88 & 1.012 & \textbf{91.40}    \\
ITC (LAION-RANDOM)   & 91.68 & 1.010  & 90.76 \\
\end{tabular}
\vspace{-1em}
\caption{Ablating the different components of FaRL pre-training w.r.t downstream task performances. \emph{ITC+MIM\textsubscript{1}+ALIGN} is the default setting of FaRL.}
\label{tab:fixbb_ablation}
\vspace{-1em}
\end{table}

We evaluate the effectiveness of different training components through ablation experiments reported in Table \ref{tab:fixbb_ablation}, where ITC represents the \emph{image text contrastive} learning, MIM stands for the \emph{masked image modeling}. MIM\textsubscript{1} means to append an additional 1-layer Transformer to $E_I$ for masked image modeling, while MIM\textsubscript{6} means to append 6 layers. ALIGN means we would follow Section \ref{ssec:pretraining_details} and randomly select one face from the original image, align and crop it to $224\times 224$ before feeding into the vision encoder. If without ALIGN, we would follow CLIP \cite{radford2021learning} instead and do random crop to get the $224 \times 224$ input. 

Compared to ITC only, adding MIM\textsubscript{1} improves performances on face parsing (LaPa) and face alignment (AFLW-19), but not on face attributes recognition (CelebA).
This supports our hypothesis that MIM helps capturing more low-level information, thus it is more beneficial for downstream tasks depending on relatively low-level features.
Although ALIGN is the most critical component for CelebA, adding both MIM\textsubscript{1} and ALIGN together achieves generally better scores on most downstream benchmarks (LaPa, AFLW-10).
In addition, a heavier MIM\textsubscript{6} head is not as good as MIM\textsubscript{1}, suggesting that the deeper head may weaken the effect of the MIM loss.
Therefore, we select ITC+MIM\textsubscript{1}+ALIGN as the default setting of FaRL.

We also append a result of an ITC-only model pre-trained on a dataset called LAION-RANDOM in Table \ref{tab:fixbb_ablation}. LAION-RANDOM has the same size with LAION-FACE but its data is randomly sampled from LAION, thus containing lots of non-face images. The model pre-trained on LAION-FACE is consistently better than the model pre-trained on LAION-RANDOM, showing the importance of face data in pre-training. The ratio of face images is more crucial for face attributes recognition (CelebA) but relatively less important for face parsing (LaPa) and face alignment (AFLW-19). It might be because face attributes recognition require reasoning on higher-level semantics. To implicitly acquire the high-level knowledge related with facial attributes and identity, the model needs to see a lot more face images in pre-training.

\subsection{Comparing with State-of-the-Art Face Methods}
\label{ssec:sota}

\begin{table}
\hspace*{-.8em}
  \centering
  \scriptsize
  \setlength\tabcolsep{1.5pt}
  \begin{tabular}{l|cccccccccc|c}

    Method & Skin & Hair & L-E & R-E & U-L & I-M & L-L & Nose & L-B & R-B & Mean \\
    \hline

    {\tiny BASS }\cite{liu2020new} & 97.2 & 96.3 & 88.1 & 88.0 & 84.4 & 87.6 & 85.7 & 95.5 & 87.7 & 87.6 & 89.8 \\
    {\tiny EHANet }\cite{luo2020ehanet} & 95.8 & 94.3 & 87.0 & 89.1 & 85.3 & 85.6 & 88.8 & 94.3 & 85.9 & 86.1 & 89.2 \\
    {\tiny Wei \etal }\cite{wei2019accurate} & 96.1 & 95.1 & 88.9 & 87.5 & 83.1 & 89.2 & 83.8 & 96.1 & 86.0 & 87.8 & 89.4 \\
    {\tiny EAGR }\cite{te2020edge} & 97.3 & 96.2 & 89.5 & 90.0 & 88.1 & 90.0 & 89.0 & 97.1 & 86.5 & 87.0 & 91.1 \\
    {\tiny AGRNet }\cite{te2021agrnet} & \underline{97.7} & \textbf{96.5} & 91.6 & 91.1 & 88.5 & \underline{90.7} & \underline{90.1} & 97.3 & 89.9 & 90.0 & 92.3 \\

    \hline
    
    Scratch & 97.18 & 93.06 & 91.61 & 91.50 & 87.22 & 89.44 & 89.13 & 97.26 & 90.12 & 89.69 & 91.62 \\ %
    FaRL & 97.38 & 94.53 & 91.88 & 91.69 & 88.20 & 90.59 & 89.85 & 97.42 & 90.84 & 90.85 & 92.32 \\
    FaRL$_\textit{ft}$ & 97.52 & 95.11 & \underline{92.33} & \underline{92.09} & \underline{88.69} & \underline{90.70} & 90.05 & \underline{97.55} & \underline{91.57} & \underline{91.34} & \underline{92.70} \\ 

    FaRL$_\textit{ft}^{448}$ & \textbf{98.00} & \textbf{96.52} & \textbf{93.97} & \textbf{93.91} & \textbf{90.15} & \textbf{91.74} & \textbf{91.21} & \textbf{97.92} & \textbf{92.70}  & \textbf{92.65}  & \textbf{93.88} \\ 

  \end{tabular}
  \vspace{-1em}
  \caption{Comparing with state-of-the-art face parsing methods on LaPa test set. Results are reported in F1 scores (\%).}
  \label{tab:face_parsing_lapa}
  \vspace{-1em}
\end{table}

\begin{table}
\hspace*{-1em}
  \centering
  \scriptsize
  \setlength\tabcolsep{2pt}
  \begin{tabular}{l|ccccccccc|c}

    \multirow{2}{*}{Method} & Face & Nose & Glasses & L-Eye & R-Eye & L-B & R-B & L-Ear & R-Ear & \multirow{2}{*}{Mean} \\
                            & I-M & U-L & L-L & Hair & Hat & Earring & Necklace & Neck & Cloth  \\
    \hline
    \multirow{2}{*}{
    \makecell{Lee \etal\\\cite{zhao2017pyramid}}} & 95.5 & 85.6 & 92.9 & 84.3 & 85.2 & 81.4 & 81.2 & 84.9 & 83.1 & \multirow{2}{*}{80.3} \\
                                                      & 63.4 & 88.9 & 90.1 & 86.6 & \underline{91.3} & 63.2 & 26.1 & \textbf{92.8} & 68.3 \\
    \hline
    \multirow{2}{*}{
    \makecell{EHANet\\\cite{luo2020ehanet}}} & 96.0 & 93.7 & 90.6 & 86.2 & 86.5 & 83.2 & 83.1 & 86.5 & 84.1 & \multirow{2}{*}{84.0} \\
                                                 & \underline{93.8} & 88.6 & 90.3 & 93.9 & 85.9 & 67.8 & 30.1 & 88.8 & 83.5 \\
    \hline
    \multirow{2}{*}{
    \makecell{Wei \etal\\\cite{wei2019accurate}}}  & \underline{96.4} & 91.9 & 89.5 & 87.1 & 85.0 & 80.8 & 82.5 & 84.1 & 83.3 & \multirow{2}{*}{82.1} \\
                                                       & 90.6 & 87.9 & 91.0 & 91.1 & 83.9 & 65.4 & 17.8 & 88.1 & 80.6 \\
    \hline
    \multirow{2}{*}{
    \makecell{EAGR\\\cite{te2020edge}}}  & 96.2 & \underline{94.0} & 92.3 & 88.6 & 88.7 & \underline{85.7} & 85.2 & 88.0 & 85.7 & \multirow{2}{*}{85.1} \\
                                             & \textbf{95.0} & 88.9 & \underline{91.2} & 94.9 & 87.6 & 68.3 & 27.6 & 89.4 & 85.3 \\
    \hline
    \multirow{2}{*}{
    \makecell{AGRNet\\\cite{te2021agrnet}}} & 96.5 & 93.9 & 91.8 & 88.7 & \underline{89.1} & 85.5 & \underline{85.6} & \underline{88.1} & \underline{88.7} & \multirow{2}{*}{85.5} \\
                                                & 92.0 & \underline{89.1} & 91.1 & 95.2 & 87.2 & 69.6 & 32.8 & 89.9 & 84.9 \\

    \hline
    
    \multirow{2}{*}{\makecell{Scratch}} & 96.17 & 93.77 & 92.28 & \underline{89.04} & 88.97 & 85.32 & 85.36 & 86.88 & 87.32 & \multirow{2}{*}{84.74} \\ 
                                        & 91.66 & 88.10 & 90.04 & 94.94 & 82.73 & 63.05 & 33.52 & 90.76 & 85.92 \\
    \hline

    \multirow{2}{*}{FaRL} & 96.29 & 93.72 & 93.91 & 88.75 & 88.64 & 85.24 & 85.42 & 87.06 & 87.36 & \multirow{2}{*}{86.72} \\ 
                                                      & 90.96 & 87.53 & 89.81 & 95.60 & 90.07 & 68.19 & 50.94 & 91.54 & 89.88 \\
    \hline

    \multirow{2}{*}{FaRL$_\textit{ft}$} & 96.32 & 93.62 & \underline{94.08} & 88.81 & 88.67 & 85.25 & 85.46 & 87.53 & 87.87 & \multirow{2}{*}{\underline{87.55}} \\ 
                             & 91.10 & 87.77 & 89.81 & \underline{95.76} & 90.80 & \underline{69.87} & \underline{60.91} & 91.79 & \underline{90.40} \\
    \hline

    \multirow{2}{*}{FaRL$_\textit{ft}^{448}$} & \textbf{96.74} & \textbf{94.22} & \textbf{95.37} & \textbf{90.71} & \textbf{90.56} & \textbf{87.03} & \textbf{87.14} & \textbf{89.06} & \textbf{89.46} & \multirow{2}{*}{\textbf{89.56}} \\ 
                                                       & 92.80   & \textbf{90.17} & \textbf{91.38} & \textbf{96.20} & \textbf{92.09} & \textbf{75.72} & \textbf{69.72} & \underline{92.45}  & \textbf{91.31}\\

  \end{tabular}
  \vspace{-1em}
  \caption{Comparing with state-of-the-art face parsing methods on CelebAMask-HQ test set. Results are reported in F1 scores (\%).}
  \label{tab:face_parsing_celebm}
  \vspace{-1em}
\end{table}

In this section, we compare FaRL with the state-of-the-art methods in multiple downstream face tasks. Different variants of FaRL are also compared. We use the name FaRL to represent our vanilla FaRL model whose pre-trained backbone is always frozen. FaRL$_\textit{ft}$ denotes the model that is fully fine-tuned from the vanilla FaRL for the specific downstream task. While both FaRL and FaRL$_\textit{ft}$ accepts $224\times 224$ inputs, we also fine-tune a model that accepts two-times larger input resolution, namely FaRL$_\textit{ft}^{448}$. FaRL$_\textit{ft}^{448}$ shares the same initial parameters with FaRL$_\textit{ft}$, but its positional embeddings are initialized by a bi-cubic up-sampling from the positional embeddings of FaRL$_\textit{ft}$. In order to investigate whether the gains are brought by our pre-training method or just come from the Transformer-based network structure, we also append a model named \emph{Scratch}. It stands for a model which shares the same network structure with FaRL, but is specifically fully trained on the corresponding downstream task from scratch.

\noindent\textbf{Face parsing}.
As illustrated in Table \ref{tab:face_parsing_lapa} and Table \ref{tab:face_parsing_celebm}, our methods achieve remarkable performances on LaPa and CelebAMask-HQ.
The vanilla FaRL surpasses the prior arts on both benchmarks. The refined FaRL$_\textit{ft}$ brings even higher F1 scores. The ultimate performance is achieved by FaRL$_\textit{ft}^{448}$, which outperforms the state-of-the-art method \cite{te2021agrnet} by 1.58 and 4.06 on LaPa and CelebAMask-HQ, respectively. We note that the input resolution plays a critical role in face parsing performance, it is especially effective for small components (\eg \emph{necklace} in CelebAMask-HQ). Even though, the required resolution of FaRL$_\textit{ft}^{448}$ is lower than the resolution needed by the state-of-the-art approach \cite{te2021agrnet} which is 473. It is also worth noting that the backbone-frozen FaRL achieves even better performances than the \emph{Scratch} models on both benchmarks, showing that the representation learned from FaRL is not only widely applicable, but also sufficiently effective.

\begin{table}
  \centering
  \scriptsize
  \begin{tabular}{l|cc|cc}

    & \multicolumn{2}{c|}{\makecell{$\mathrm{NME}_\mathrm{diag}$ $\downarrow$}} 
    & \makecell{$\mathrm{NME}_\mathrm{box}$ $\downarrow$} 
    & \makecell{$\mathrm{AUC}_\mathrm{box}^7$ $\uparrow$} \\

    Method & Full  & Frontal & Full & Full \\
    \hline
    CFSS \cite{zhu2015face}  & 3.92 & 2.68 & - & -  \\
    CCL \cite{zhu2016unconstrained} & 2.72 & 2.17 & - & - \\
    DAC-CSR \cite{feng2017dynamic} & 2.27 & 1.81 & - & - \\
    LLL \cite{robinson2019laplace} & 1.97 & - & - \\
    SAN \cite{dong2018style} & 1.91 & 1.85 & 4.04 & 54.0 \\
    DSRN \cite{miao2018direct} & 1.86 & - & - \\
    LAB (w/o B) \cite{wu2018look} & 1.85 & 1.62 & - & - \\
    LAB (w/ B) \cite{wu2018look} & 1.25 & 1.14 & - & - \\
    HR-Net \cite{sun2019high} & 1.57 & 1.46 & - & - \\
    Wing \cite{feng2018wing} & - & - & 3.56 & 53.5 \\
    KDN \cite{chen2019face} & - & - & 2.80 & 60.3 \\
    LUVLi \cite{kumar2020luvli} & 1.39 & 1.19 & 2.28 & 68.0 \\

    Bulat \etal \cite{bulat2021pre} & 1.54 & - & - & - \\

    \hline

    Scratch    & 1.047 & 0.884  & 1.481 & 79.3 \\ 
    FaRL               &  0.991 & 0.851 & 1.402 & 80.4 \\ 
    FaRL$_\textit{ft}$ & \underline{0.969}  & \underline{0.836} & \underline{1.371} & \underline{80.8} \\ 
    FaRL$_\textit{ft}^{448}$ & \textbf{0.943} & \textbf{0.821} & \textbf{1.334} & \textbf{81.3}  \\ 

  \end{tabular}
  \vspace{-1em}
  \caption{Comparing with state-of-the-art face alignment methods on two AFLW-19 test sets: the Full set and the Frontal subset.}
  \label{tab:face_landmark_aflw-19}
  \vspace{-1em}
\end{table}

\begin{table}
  \centering
  \scriptsize
  \setlength\tabcolsep{1pt}
  \begin{tabular}{l|c|ccc ccc|cc}
    
     & \multicolumn{7}{c|}{$\mathrm{NME}_\mathrm{inter\textit{-}ocular} \downarrow$}
     & $\mathrm{FR}^{10} \downarrow$ 
     & $\mathrm{AUC}^{10} \uparrow$ \\


    Method   & Full  & Pose & Expr. & Illum. & MakeUp & Occl. & Blur & \multicolumn{2}{c}{Full} \\
    \hline
    
    ESR \cite{cao2014face} & 11.13 & 25.88 & 11.47 & 10.49 & 11.05 & 13.75 & 12.20   & 35.24 & 27.74 \\
    SDM \cite{xiong2013supervised} & 10.29 & 24.10 & 11.45 & 9.32 & 9.38 & 13.03 & 11.28     & 29.40 & 30.02 \\
    CFSS \cite{zhu2015face} & 9.07 & 21.36 & 10.09 & 8.30 & 8.74 & 11.76 & 9.96      & 20.56 & 36.59 \\
    DVLN \cite{wu2017leveraging} & 6.08 & 11.54 & 6.78 & 5.73 & 5.98 & 7.33 & 6.88        & 10.84 & 45.51 \\
    LAB \cite{wu2018look} & 5.27 & 10.24 & 5.51 & 5.23 & 5.15 & 6.79 & 6.12         & 7.56  & 53.23 \\
    Wing \cite{feng2018wing} & 5.11 & 8.75 & 5.36 & 4.93 & 5.41 & 6.37 & 5.81         & 6.00  & 55.04 \\
    DeCaFA \cite{dapogny2019decafa} & 4.62 & 8.11 & 4.65 & 4.41 & 4.63 & 5.74 & 5.38       & 4.84  & 56.30 \\
    AWing \cite{wang2019adaptive} & 4.36 & 7.38 & 4.58 & 4.32 & 4.27 & 5.19 & 4.96        & 2.84  & 57.19 \\
    LUVLi \cite{kumar2020luvli} & 4.37 & - & - & - & - & - & -                       & 3.12  & 57.70 \\
    ADNet \cite{huang2021adnet} & 4.14 & 6.96 & 4.38 & 4.09 & 4.05 & 5.06 & 4.79   & 2.72  & 60.22 \\

    Bulat \etal \cite{bulat2021pre} & 4.57 & - & - & - & - & - & - & - & - \\

    \hline

    Scratch & 4.80 & 8.78 & 5.09 & 4.74 & 4.99 & 6.01 & 5.35 & 5.72 & 54.54\\ 
    FaRL  &  4.38 & 7.60 & 4.66 & 4.19 & 4.30 & 5.44 & 4.98 & 3.32 & 57.54 \\ 
    FaRL$_\textit{ft}$   & \underline{4.03} & \textbf{6.81} & \underline{4.32} & \textbf{3.92} & \underline{3.87} & \textbf{4.70} & \textbf{4.54} & \textbf{1.76} & \underline{60.23} \\ 
    FaRL$_\textit{ft}^{448}$ & \textbf{3.96} & \underline{6.91} & \textbf{4.21} & \underline{3.97} & \textbf{3.80} & \underline{4.71} & \underline{4.57} & \textbf{1.76} & \textbf{61.16} \\ 

  \end{tabular}
  \vspace{-1em}
  \caption{Comparing with state-of-the-art face alignment methods on WFLW test sets. }
  \label{tab:face_landmark_wflw}
  \vspace{-1em}
\end{table}

\begin{table}
  \centering
  \scriptsize
  \begin{tabular}{l|cc|c}
           & \multicolumn{3}{c}{$\mathrm{NME}_\mathrm{inter\textit{-}ocular}$ $\downarrow$}  \\
    Method & Common  & Challenge & Full \\
    \hline
    SAN \cite{dong2018style}       & 3.34 & 6.60 & 3.98 \\
    AVS \cite{qian2019aggregation} & 3.21 & 6.49 & 3.86 \\
    DAN \cite{kowalski2017deep}    & 3.19 & 5.24 & 3.59 \\
    LAB (w/ B) \cite{wu2018look}   & 2.98 & 5.19 & 3.49 \\
    DCFE (w/ 3D) \cite{valle2018deeply} &  2.76 & 5.22 & 3.24 \\
    Teacher \cite{dong2019teacher} & 2.91 & 5.91 & 3.49 \\
    DU-Net \cite{tang2019towards} & 2.97 & 5.53 & 3.47 \\
    DeCaFa \cite{dapogny2019decafa} & 2.93 & 5.26 & 3.39 \\
    HR-Net \cite{sun2019high} & 2.87 & 5.15 & 3.32 \\
    HG-HSLE \cite{zou2019learning} & 2.85 & 5.03 & 3.28 \\
    AWing \cite{wang2019adaptive} & 2.72 & 4.52 & 3.07 \\
    LUVLi \cite{kumar2020luvli} & 2.76 & 5.16 & 3.23 \\
    ADNet \cite{huang2021adnet} & \textbf{2.53} & \underline{4.58} & \textbf{2.93} \\

    Bulat \etal \cite{bulat2021pre} & - & - & 3.85 \\

    \hline
    Scratch & 2.90 & 5.19 & 3.35 \\ 
    FaRL & 2.69 & 4.85 & 3.12 \\ 
    FaRL$_\textit{ft}$  & 2.70 & 4.64 & 3.08 \\ 
    FaRL$_\textit{ft}^{448}$ & \underline{2.56} & \textbf{4.45} & \textbf{2.93} \\ 

  \end{tabular}
  \vspace{-1em}
  \caption{Comparing with state-of-the-art face alignment methods on three 300W test sets: the Common subset, the Challenge subset and the Full set. }
  \label{tab:face_landmark_300w}
  \vspace{-2em}
\end{table}

\noindent\textbf{Face alignment}. We compare with previous face alignment methods on three benchmarks: AFLW-19, 300W and WFLW, and report results in Table \ref{tab:face_landmark_aflw-19}, Table \ref{tab:face_landmark_300w} and Table \ref{tab:face_landmark_wflw}, respectively. The Transformer-based methods achieves performances superior to all prior arts on AFLW-19, as shown by Table \ref{tab:face_landmark_aflw-19}. Among these Transformer-based methods, the vanilla FaRL consistently outperforms \emph{Scratch}, and our FaRL$_{ft}^{448}$ achieves new state-of-the-art performances on both AFLW-19 and WFLW, while being comparable with \cite{huang2021adnet} on 300W. But unlike \cite{huang2021adnet}, our method does not assume any co-boundary relationship among landmark points. In addition, our methods outperform a previous work \cite{bulat2021pre}, which also leverages unified face representation pre-training, by a large margin.

\begin{table}
  \centering
  \scriptsize
  \setlength\tabcolsep{1pt}
  \begin{tabular}{l|cccc c|ccc cc}
   & \multicolumn{5}{c|}{CelebA}
   & \multicolumn{5}{c}{LFWA} \\
  \hline
  Proportion & $0.2\%$ & $0.5\%$ & $1\%$ & $2\%$ & $100\%$ & $5\%$ & $10\%$ & $20\%$ & $50\%$ & $100\%$ \\
  \# of training data & 325 & 843 & 1,627 & 3,255 & 162,770 & 313 & 626 & 1,252 & 3,131 & 6,263 \\
  \hline
  DMM \cite{mao2020deep}  & - & - & - & - & 91.70 & - & - & -  & - & 86.56 \\
  SlimCNN \cite{sharma2020slim} & 79.90 & 80.20 & 80.96 & 82.32 & 91.24 & 70.90 & 71.49 & 72.12  & 73.45 & 76.02\\
  AFFAIR \cite{li2018landmark} & - & - & - & - & 91.45 & - & - & -  & - & 86.13 \\
  PS-MCNN \cite{cao2018partially} & - & - & - & - & \textbf{92.98} & - & - & -  & - & \textbf{87.36} \\
  He \etal \cite{he2018harnessing} & - & - & - & - & 91.81 & - & - & -  & - & 85.20 \\

  DeepCluster \cite{caron2018deep} & 83.21& 86.13& 87.46& 88.86& 91.68& 74.21& 77.42& 80.77& 84.27& 85.90\\
  JigsawPuzzle \cite{noroozi2016unsupervised} & 82.88& 84.71& 86.25& 87.77& 91.57& 73.90& 77.01 & 79.56 & 83.29 & 84.86\\
  Rot \cite{gidaris2018unsupervised} & 83.25 & 86.51 & 87.67& 88.82&91.69&74.40&76.67&81.52&84.90&85.72\\
  FixMath \cite{sohn2020fixmatch} & 80.22 & 84.19 & 85.77 & 86.14 & 89.78 & 71.42 & 72.78 & 75.10 & 80.87 & 83.84 \\
  VAT \cite{miyato2018virtual} & 81.44 & 84.02 & 86.30 & 87.28 & 91.44 & 72.19 & 74.42 & 76.26 & 80.55 & 84.68\\

  SSPL \cite{shu2021learning} & 86.67 & 88.05 & 88.84 & 89.58 & 91.77 & 78.68 & 81.65 & 83.45 & \underline{85.43} & 86.53 \\
  
  \hline
  
  FaRL & \underline{87.63} & \underline{88.58} & \underline{89.66} & \underline{90.26} & 91.39 & \underline{80.94} & \underline{82.42} & \underline{83.94} & 85.38 & 85.94\\
  FaRL$_\textit{ft}$ & \textbf{88.51} & \textbf{89.12}  & \textbf{90.24} & \textbf{90.55} & \underline{91.88} & \textbf{82.57} & \textbf{83.58} & \textbf{84.80} & \textbf{85.95} & \underline{86.69} \\

  \end{tabular}
  \vspace{-1em}
  \caption{Comparing with other face attributes recognition methods under both full-shot and few-shot protocols on CelebA and LFWA. Results are reported in mean accuracy (\%).}
  \label{tab:face_attrs_fewshot}

\end{table}

\noindent\textbf{Face attributes recognition}. We compare with previous methods under both full-shot and few-shot settings on two benchmarks: CelebA and LFWA. As shown in Table \ref{tab:face_attrs_fewshot}, our method outperforms others reported under all few-shot settings, while ranking the second under full-shot. A possible reason is that the state-of-the-art method \cite{cao2018partially} explicitly leverages extra information from downstream data, including face identity annotations and hand-designed attribute relationships.

\section{Discussions}

In this section, we discuss the pros and cons of our model.
While some of these have been analyzed in previous various sections,
we summarize and collect them here.

\noindent 
\textbf{The benefits of our model.}
First and most important, our model provides a general facial representation learning framework showing excellent performance on downstream tasks. In this way, a face image can be fed into a general feature extraction module once and for all, then the feature is used in different decision modules for different respective face tasks. This is particularly useful for resource-restricted mobile devices.
Second, our model is applicable for low-level face tasks as well as high-level face tasks, due to the joint learning of the contrastive learning capturing the semantics and the masked image modeling harnessing the low-level information.
Last, we show that the framework achieves satisfactory results based on 20$M$ samples, much fewer compared with general image representation pre-training \cite{radford2021learning} utilizing hundreds of millions of samples. This will facilitate its adoption.

\noindent
\textbf{Ethical considerations.}
Our model relies on the large-scale image-text data which is usually crawled from the Internet with a huge amount of data publicly available in such form.
While crawling face images seems to be another matter. It may suffer from social biases as it would be unlikely to fully cover all different aspects of face-related queries when crawling.
Although we choose a roundabout way of filtering out face images from a general image-text pair dataset, it is still possible that bias exists in the extracted subdataset.

Thereupon we provide preliminary analysis on the face image dataset FairFace~\cite{karkkainenfairface}, FairFace is a face attribute dataset for balanced race, gender, and age. It categorizes gender into 2 groups, race into 7 groups, age into 9 groups. 
We freeze the pretrained backbones and only fine-tune classification heads on FairFace.
We compare FaRL with the 400M-data-pretrained CLIP \cite{radford2021learning}\footnote{We compare with CLIP-ViT-B/16 under the same fine-tuning protocol for fairness.}, as well as the baseline model associated with FairFace \cite{karkkainenfairface}.
The accuracies w.r.t race groups are reported in Table \ref{tab:fairface}. We follow \cite{radford2021learning} and define the group ``Non-White'' to include multiple race categories: ``Black'', ``Indian'', ``East Asian'', ``Southeast Asian'', ``Middle Eastern'' and ``Latino''.

We are interested in the performance gaps between different race groups, which might associate with social bias. As shown by the right-most column of Table \ref{tab:fairface}, though FaRL exhibits gaps between different race groups, the gap values of FaRL are relatively moderate among others.

\begin{table}
\centering
\setlength\tabcolsep{2pt}
\footnotesize
\begin{tabular}{c|l|cc|cc}
                        & Model    & White & Non-White & Average   & Discrepancy  \\
\hline
\multirow{3}{*}{Age}    & FairFace~\cite{karkkainenfairface} & 60.05 & 60.63    & 60.52   & +0.58 \\
                        & CLIP~\cite{radford2021learning}     & {62.25} & {61.95}    & {62.00}  & {-0.30} \\
                        & FaRL      & {61.49} & {61.84}    & {61.78}   & +0.35 \\
\hline
\multirow{3}{*}{Gender} & FairFace~\cite{karkkainenfairface} & 94.15 & 94.41    & 94.36   & {+0.26} \\
                        & CLIP~\cite{radford2021learning}     & {94.87} & {95.78}    & 95.61  & +0.91 \\
                        & FaRL      & {95.16} & {95.77}    & {95.65}   & +0.61 \\
\end{tabular}
\vspace{-1em}
\caption{Age and gender classification accuracies on FairFace. Results are reported w.r.t two race groups.}
\label{tab:fairface}
\end{table}

\noindent
\textbf{Limitations.} We are aware that there are still limitations of our model.
First, as discussed above, our model presents bias to a certain degree,
which might be caused by: 1) the data bias existing in the original LAION dataset \cite{schuhmann2021laion}, or 2) the performance bias of the face detector \cite{deng2020retinaface} we use.
Second, our current work has not yet adapted to some important face tasks, \eg, face detection, face anti-spoofing and face forgery detection. 
We expect our future updates would address these issues.

\section{Conclusion}

In this paper, we investigate the transfer performance of pre-trained models on face analysis tasks. 
We design a pre-training method, called FaRL, that leverages image-text contrastive learning as well as masked image modeling, to learn more general facial representation.
We show that the face representation learned by FaRL transfers well to downstream face analysis tasks, including face parsing, face alignment, and face attributes recognition. 
Compared with previous pre-trained models, our model FaRL achieves superior transfer performance. Moreover, the proposed model surpasses the state-of-the-art methods on face parsing and face alignment.
 
\vspace{0.5em}
\noindent\textbf{Acknowledgements}
Yinglin Zheng and Ming Zeng were partially supported by NSFC (No. 62072382), Fundamental Research Funds for Central Universities, China (No. 20720190003).

\appendix

\section{Evaluation on Face Editing Tasks.}
Here we adopt a recent text-driven face editing framework \cite{patashnik2021styleclip}, which uses a pre-trained CLIP for visual-language reasoning.
We replace CLIP with our FaRL (with equal model size), and show comparisons below.
It can be observed that, the generated face images which are driven by FaRL are more faithful to the given text prompts.

\vspace{-1em}
\begin{figure}[h]
  \centering
  \includegraphics[width=1.0\linewidth]{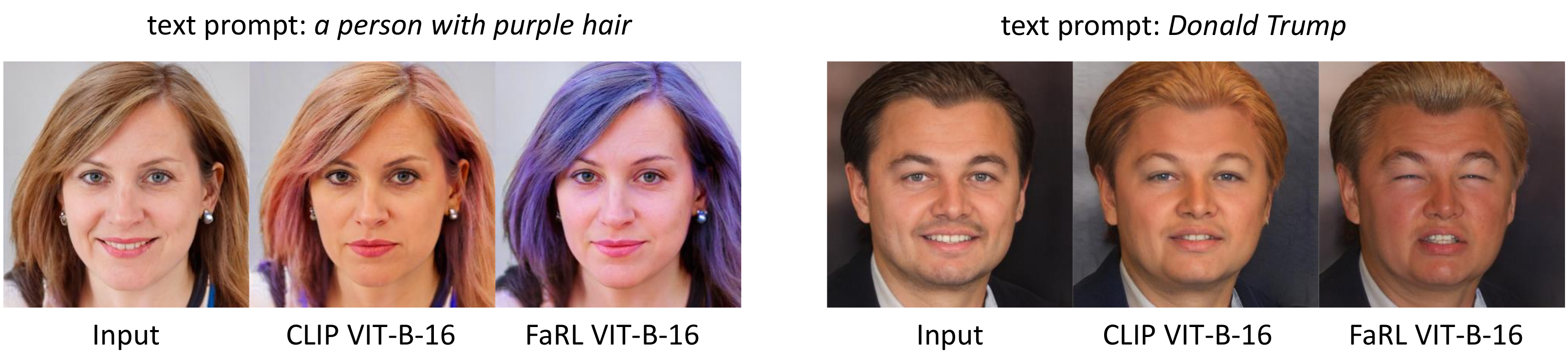}
    \vspace{-2em}
  \caption{Comparing FaRL with CLIP in text-driven face editing.
  }
  \label{fig:stylefarl}

\end{figure}
\vspace{-1em}

\section{Visualizing the Pre-trained Image Encoder}

\begin{figure}[t]
  \centering
  \includegraphics[width=1.0\linewidth]{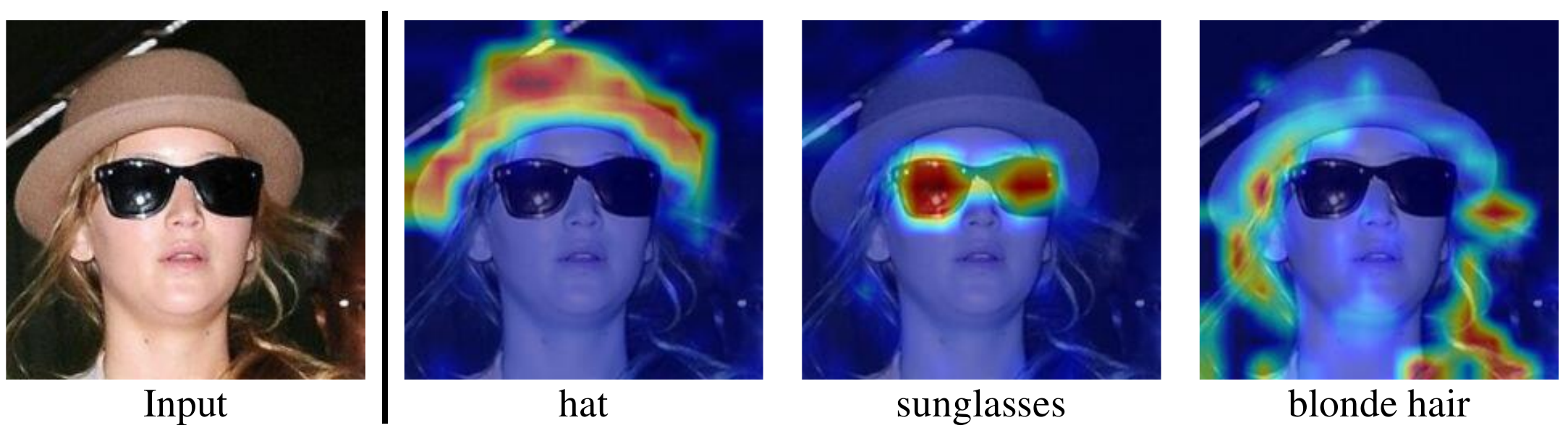}
  \caption{Grad-CAM visualizations of $E_I$ given different text queries. Gradients are calculated in output of the first LayerNorm within the last Transformer block of $E_I$.}
  \label{fig:gradcam_image}
\end{figure}

In Figure \ref{fig:gradcam_image}, we provide the Grad-CAM \cite{selvaraju2017grad} visualizations for the pre-trained FaRL image encoder $E_I$, with different text queries fed into the text encoder $E_T$. Gradients are calculated in the output of the first LayerNorm within the last Transformer block of $E_I$.
As can be seen in the figure, 
our image encoder successfully localizes the corresponding regions for different query texts, showing a high correlation with human attention.

\section{Features on Different Backbone Levels}

\begin{figure}
  \centering
  \begin{subfigure}[t]{\linewidth}
  \includegraphics[width=1.0\linewidth]{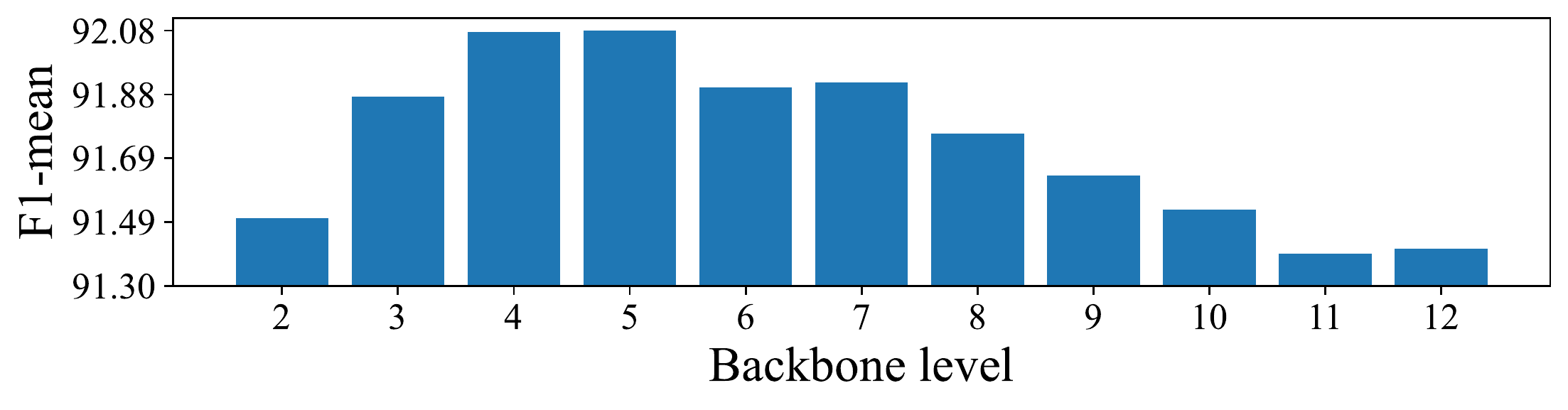}
  \caption{LaPa (face parsing).}
  \label{fig:lapa_layers}
  \end{subfigure}
  \begin{subfigure}[t]{\linewidth}
  \centering
  \includegraphics[width=1.0\linewidth]{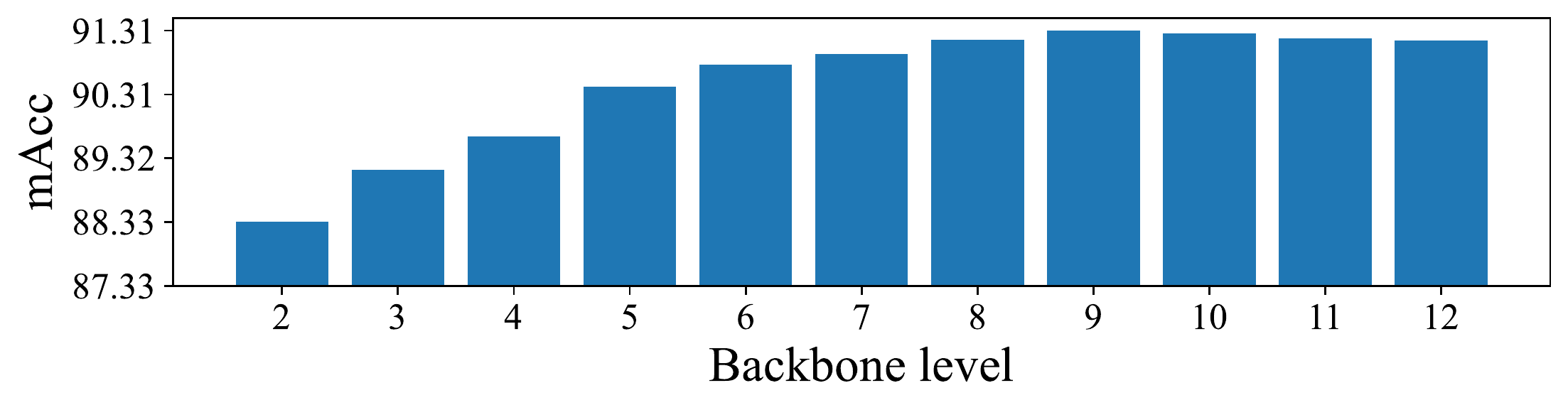}
  \caption{CelebA (face attributes recognition).}
  \label{fig:celeb_layers}
  \end{subfigure}
  \caption{Performance of features from different backbone levels on face parsing (LaPa) and face attributes recognition (CelebA).}
  \label{fig:layers}
\end{figure}

Instead of integrating multi-level features in downstream tasks, we also study how features on each single level of $E_I$ affect the performances. We replace the multi-level features with repeated single feature on each level and apply the same head for downstream evaluation with backbone frozen. Figure \ref{fig:lapa_layers} and Figure \ref{fig:celeb_layers} illustrate the corresponding performances on LaPa (face parsing) and CelebA (face attributes recognition), respectively. In these two figures, a larger backbone level number indicates a deeper feature.

It is interesting to see that features on different backbone levels behave in quite divergent ways for different downstream tasks. The features on deep levels (e.g level 9) are the most effective for face attributes recognition on CelebA. However, they perform poorly on face parsing: the most effective feature for face parsing is on the 5-th level instead. This suggests that 1) the encoder $E_I$ has learned different kinds of semantics on different feature levels during pre-training; 2) different kinds of downstream tasks require different kinds of semantics. 
Tasks like face attributes recognition rely more on high-level semantics, while face parsing is more in favor of low-level ones. In consideration of this divergence, it 
might not be the best way to always use the feature from just one single level for all downstream tasks.  

Besides, we observe that the fusion of multi-level features, which is adopted by FaRL, achieves 92.32 (F1-mean) on LaPa and 91.39 (mAcc) on CelebA, outperforming all single-level settings. This indicates a complementary nature among features on different backbone levels.

\section{Ratios of Face Images in Pre-training}

We are also curious about how the ratio of face images used in pre-training affects the performances of downstream face tasks.
Therefore, we randomly sample different numbers of face images from LAION \cite{schuhmann2021laion} and construct different pre-training datasets for investigation.
All these datasets share the same size with LAION-FACE (20M), but with different ratios of face images within.
We conduct pre-training on these datasets using only image-text contrastive learning.

The corresponding performances of different downstream face tasks are illustrated in Table \ref{tab:face_ratio}. 
In general, a higher ratio of face images leads to a better performance. The gains are significant on tasks like face attributes recognition, but are quite subtle on tasks like face parsing and face alignment.

Such observation matches our previous hypothesis, that different downstream tasks require different kinds of semantics. Those high-level facial semantics, which are necessary for face attributes recognition, can only be learned on face images; while those low-level semantics (\eg corners, edges), which are required by face parsing and face alignment, can be learned from not only face images, but also non-face images as well. 
This also explains why the advantages of our FaRL shown on face parsing and face alignment over those Transformers pre-trained on general images (\eg ImageNet, WIT), are not as large as the advantages exhibited on face attributes recognition.

\begin{table}[t]
\centering
\footnotesize
\setlength\tabcolsep{4pt}
\begin{tabular}{c|ccc}

\makecell{\% of\\Face Images}
& \makecell{LaPa\\F1-mean$\uparrow$} & \makecell{AFLW-19\\$\mathrm{NME}_\mathrm{diag}\downarrow$} 
& \makecell{CelebA\\mAcc$\uparrow$}  \\
\hline

0 & 91.68 & 1.017 & 89.73 \\
12.5 & 91.68 & 1.010 & 90.76 \\
50 & \textbf{91.77} & \textbf{1.009} & 91.17 \\
100 & 91.75 & \textbf{1.009} & \textbf{91.31} 
\\

\end{tabular}
\caption{Downstream performances w.r.t different face image ratios in pre-training data. Here the pre-training only uses image-text contrastive learning.}
\label{tab:face_ratio}
\end{table}

\section{Comparison with Self-supervised Methods on Face Dataset}
Here we compare FaRL with two recent self-supervised methods: SwAV\cite{caron2020unsupervised} and SimCLR\cite{chen2020simple} on LAION-Face with the same network structure and fine-tuning strategy with FaRL. Note that SwAV on face images is equivalent to Bulat\cite{bulat2021pre}.
as it adopts SwAV for face pre-training.
As shown in Table \ref{tab:compare_pretrain}, FaRL achieves better performances on all tasks.

\begin{table}[h]
\centering
\footnotesize
\setlength\tabcolsep{2pt}
\begin{tabular}{r|ccc}
\makecell{Pre-training Settings} 

& \makecell{LaPa\\F1-mean$\uparrow$} & \makecell{AFLW-19\\$\mathrm{NME}_\mathrm{diag}\downarrow$} 
& \makecell{CelebA\\mAcc$\uparrow$}  \\
\hline 

(Bulat\cite{bulat2021pre}) SwAV+ALIGN & 90.55 & 1.059 & 89.65\\
SimCLR+ALIGN & 91.72 & 0.995 & 91.08 \\
\hline
(FaRL) {ITC+MIM\textsubscript{1}+ALIGN}         & \textbf{92.32} & \textbf{0.991}   & \textbf{91.39}\\

\end{tabular}
\caption{Comparison with self-supervised pre-training on face data.
}
\label{tab:compare_pretrain}
\end{table}

\section{Longer Pre-training for Finetuning}

We illustrate finetuning results from longer FaRL pre-training in Table \ref{tab:longer_farl_pre-training}. Additional benefits can be observed from longer pre-training.

\begin{table*}[t]
\centering
\footnotesize
\setlength\tabcolsep{2pt}
\begin{tabular}{c|c|cc|ccc}
& \makecell{Pre-training Epoches}
& \makecell{LaPa\\F1-mean$\uparrow$} 
& \makecell{CelebAMask-HQ\\F1-mean$\uparrow$}
& \makecell{AFLW-19\\$\mathrm{NME}_\mathrm{diag}\downarrow$} 
& \makecell{WFLW\\$\mathrm{NME}_\mathrm{inter\textit{-}ocular}\downarrow$}
& \makecell{300W\\$\mathrm{NME}_\mathrm{inter\textit{-}ocular}\downarrow$} \\
\hline 
\multirow{2}{*}{\makecell{FaRL$_{ft}^{448}$}} & 
16   &  93.88 & 89.56 & 0.943 & 3.96 & 2.93  \\
& 64 &  \textbf{94.04} &\textbf{89.57} & \textbf{0.938} & \textbf{3.88} & \textbf{2.88}  \\
\end{tabular}
\caption{Finetuning results of longer FaRL pre-training.
}
\label{tab:longer_farl_pre-training}
\end{table*}

\section{More Details}

\noindent\textbf{Pre-training.} During training, mixed-precision was used to accelerate training and save memory. 
Gradient checkpointing and ZeRO\cite{rajbhandari2020zero} are also used for further memory efficiency. Gradient clip with max norm of 1.0 is applied to stabilize the training process.
Our implementation of image-text contrastive learning differs from CLIP \cite{radford2021learning}, which computes contrastive loss using only the local batch on each GPU, our implementation gathers all logits from all GPUs and consider all of them in contrastive learning. 

\noindent\textbf{Computational Complexity.} With the same image encoder structure (ViT-B), the computational complexity of our model is exactly the same with other pre-training methods during both downstream training and downstream inference. While during pre-training, 
our model has an extra text encoder and an additional MIM stage, leading to a generally doubled computation complexity comparing with CLIP; but we share comparable computation complexity with self-supervised contrastive learning methods (e.g. MoCo v3, SimCLR).

\noindent\textbf{Face parsing}. We adopt augmentations to face parsing tasks.
On LaPa, we first compute a face alignment matrix that aligns five face landmarks retrieved from a face detector \cite{deng2020retinaface} to the landmarks of a mean face in a target resolution $s \in \{224, 448\}$.
We then augment on the matrix with random rotation within $[-18^\circ, 18^\circ]$, random rescaling within $[0.9, 1.1]$ and random translation with a range of $0.01\times s$.
We transform both the image and the groundtruth label maps using the augmented matrix. 
The transformation is combined with the Tanh-warping \cite{lin2019face} to ensure that the network can segment the whole face image as well as focusing on the face region. In order to better preserve linearity within face region, we modify the warping function of \cite{lin2019face} from $\tanh$ to $\tanh_{\alpha}$ defined as $\tanh_{\alpha}(x)=$
\begin{align}
 \left\{ 
	\begin{array}{cl}
		x, & -1+\alpha \leq x \leq 1-\alpha \\
		\alpha\tanh\left(\frac{x-1+\alpha}{\alpha}\right)+1-\alpha, & 1-\alpha<x\\
		\alpha\tanh\left(\frac{x+1-\alpha}{\alpha}\right)-1+\alpha, & x<-1+\alpha\\
	\end{array}
	\right.,\nonumber
\end{align}
with $\alpha$ being a warping factor: $\tanh_{\alpha=1.0}$ equals to a vanilla $\tanh$ warping, while $\tanh_{\alpha\rightarrow 0.0}$ degenerates to a crop function that drops all peripheral pixels. 
Table \ref{tab:face_parsing_tanh_alpha} show results under different $\alpha$. $\alpha=0.8$ is selected as the default setting.

\begin{table}[t]
\centering
\scriptsize
\setlength\tabcolsep{4pt}
\begin{tabular}{l|c|c|c}
& \multicolumn{3}{c}{F1 $\uparrow$} \\
$\alpha$
& Mean & Mean (w/o Hair) & Hair  \\
\hline
0.0 & 91.51 & 91.93 & 87.74 \\ 
0.2 & 92.08 & 92.04 & 92.52 \\ 
0.4 & 92.27 & 92.07 & 94.09 \\ 
0.6 & 92.31 & 92.07 & \textbf{94.54}\\ 
0.8 (default) & \textbf{92.32} & \textbf{92.08} & 94.53 \\ 
1.0 & 92.11 & 91.84 & 94.49 \\ 

\end{tabular}
\caption{F1 scores on LaPa under different warping factors.}
\label{tab:face_parsing_tanh_alpha}
\end{table}

On CelebAMask-HQ, we replace the face alignment matrix to be a simple rescaling matrix that resizes the original size to $s \times s$, since all face images in CelebAMask-HQ are already aligned. We also disable Tanh-warping. The rest augmentations all remain the same with those on LaPa. All these above setups are also adopted for all other pre-trained models for fair comparison.

\noindent\textbf{Face alignment}. Augmentations are also applied to face alignment tasks. Random geometric transformations are first applied on the bounding boxes provided by the corresponding face alignment datasets. These transformations include random rotation within $[-10^\circ, 10^\circ]$, random rescaling within $[0.9, 1.1]$ and random translation with a range of $0.01\times s$. 
The same transformations are imposed on the groundtruth 2D face landmarks as well.
Then, we crop the original image with the transformed bounding boxes and rectify the groundtruth landmark coordinates accordingly. 
Finally, all these augmented groundtruth landmark points are rendered to $128\times 128$ heatmaps for training, using an on-line approach. 
Random Gaussian blur, noise and occlusion are also used on the input images.
These setups are also adopted for all other pre-trained models for fair comparison.

\noindent\textbf{Face attributes recognition}. Since our model is pretrained with aligned face images, it's important to also train the downstream task with aligned images. For CelebA dataset, we use the facial landmarks delivered with the dataset, for LFWA dataset, we do face detection with RetinaFace\cite{deng2020retinaface}. When training with backbone frozen, we randomly horizontal-flip the image with a probability of 0.5. When fully fine-tuning the model, apart from the random horizontal flip, random crop and Gaussian noise, we also apply random grayscale with a probability of 0.1, and impose Gaussian noise with a variance of 5 to the facial landmarks used for aligning the face. These setups are also adopted for all other pre-trained models for fair comparison.

\section{Data Usage}

\noindent\textbf{LAION} \cite{schuhmann2021laion}\footnote{\url{https://laion.ai/laion-400-open-dataset/}} contains 400M image-text pairs that are collected from Internet. It is licensed under Creative Common CC-BY 4.0. They don't claim copyright of the images.

\noindent\textbf{LaPa} \cite{liu2020new}\footnote{\url{https://github.com/JDAI-CV/lapa-dataset}} contains over 22K face images. Its license says
``\emph{this LaPa Dataset is made freely available to academic and non-academic entities for non-commercial purposes such as academic research, teaching, scientific publications, or personal experimentation. Permission is granted to use the data given that you agree to our license terms}''.

\noindent\textbf{CelebAMask-HQ} \cite{lee2020maskgan}\footnote{\url{https://github.com/switchablenorms/CelebAMask-HQ}} contains 30K face images. The usage of the dataset is {restricted to non-commercial research and educational purposes}.

\noindent\textbf{AFLW-19} \cite{zhu2016unconstrained}\footnote{\url{http://mmlab.ie.cuhk.edu.hk/projects/compositional.html}} contains around 24K face images. It is revised from the original AFLW dataset \cite{koestinger2011annotated}\footnote{\url{https://www.tugraz.at/institute/icg/research/team-bischof/lrs/downloads/aflw/\#license}} without any claims on licensing. The original AFLW dataset is available for non-commercial research purposes only.

\noindent\textbf{WFLW} \cite{wu2018look}\footnote{\url{https://wywu.github.io/projects/LAB/WFLW.html}} contains images of about 10K faces. It does not mention any licenses.

\noindent\textbf{300W} \cite{sagonas2016300,sagonas2013semi,sagonas2013300}\footnote{\url{https://ibug.doc.ic.ac.uk/resources/300-W/}} contains over 4K face images. The data are provided for research purposes only. Commercial use (\ie, use in training commercial algorithms) is not allowed.

\noindent\textbf{CelebA \& LFWA} \cite{liu2015deep}\footnote{\url{https://mmlab.ie.cuhk.edu.hk/projects/CelebA.html}}. CelebA has 202,599 face images while LFWA has 13,143. The CelebA dataset is available for non-commercial research purposes only. The LFWA dataset is based on the original LFW dataset \cite{LFWTech}\footnote{\url{http://vis-www.cs.umass.edu/lfw/}}.

\section{Code Usage}

\noindent\textbf{ViT} \cite{dosovitskiy2020image}, \noindent\textbf{DeiT} \cite{touvron2021training}. 
We use the \texttt{timm} library\footnote{\url{https://github.com/rwightman/pytorch-image-models}} to load these two pre-trained Transformers. 
We load the ViT model with \texttt{vit\_base\_patch16\_224\_in21k}, and load DeiT with \texttt{deit\_base\_distilled\_patch16\_224}.
The code of \texttt{timm} is licensed under Apache 2.0. 
Please refer to its \href{https://github.com/rwightman/pytorch-image-models#pretrained-weights}{website} for the licensing of its pre-trained weights.

\noindent\textbf{MoCo v3} \cite{chen2021empirical}\footnote{\url{https://github.com/facebookresearch/moco-v3}}. We download its \href{https://dl.fbaipublicfiles.com/moco-v3/vit-b-300ep/vit-b-300ep.pth.tar}{ViT-Base model} and use the provided script to convert the weights to DeiT format, which is then loaded by the \texttt{timm} library. MoCo v3 is under the CC-BY-NC 4.0 license. 

\noindent\textbf{BEiT} \cite{chen2021empirical}\footnote{\url{https://github.com/microsoft/unilm/tree/master/beit}} is under MIT License. Its \href{https://unilm.blob.core.windows.net/beit/beit_base_patch16_224_pt22k.pth}{BEiT-base} is used.

\noindent\textbf{CLIP} \cite{radford2021learning}\footnote{\url{https://github.com/openai/CLIP}} is under MIT license. Its ViT-B/16 is used.

\noindent\textbf{FaceTransformer}\cite{zhong2021face}%
\footnote{\url{https://github.com/zhongyy/Face-Transformer}} does not contain a license. Its \href{https://github.com/zhongyy/Face-Transformer#4-pretrained-models-and-test-models-on-lfw-sllfw-calfw-cplfw-talfw-cfp_fp-agedb}{ViT-P8S8} is used.

\noindent\textbf{RetinaFace} \cite{deng2020retinaface}\footnote{\url{https://github.com/biubug6/Pytorch_Retinaface}} is used for face detection. It is under the MIT license.

\noindent\textbf{MMSegmentation}\footnote{\url{https://github.com/open-mmlab/mmsegmentation}} is under the Apache 2.0 license. We use its UperNet \cite{xiao2018unified} implementation for downstream tasks like face parsing and face alignment.

{\small
\bibliographystyle{ieee_fullname}
\bibliography{egbib}
}

\end{document}